\documentclass{article}

\usepackage{neurips_2022}

\usepackage[utf8]{inputenc} 
\usepackage[T1]{fontenc}    
\usepackage{hyperref}       
\usepackage{url}            
\usepackage{booktabs}       
\usepackage{amsfonts}       
\usepackage{nicefrac}       
\usepackage{microtype}      
\usepackage{xcolor}         
\usepackage[thinc]{esdiff}
\usepackage{amsmath}
\usepackage{graphicx}

\title{A comparative study of back propagation and its alternatives on multilayer perceptrons}

\begin{document}

\maketitle

\begin{abstract}
The de facto algorithm for training the back pass of a feedforward neural network is backpropagation (BP). The use of almost-everywhere differentiable activation functions made it efficient and effective to propagate the gradient backwards through layers of deep neural networks. However, in recent years, there has been much research in alternatives to backpropagation. This analysis has largely focused on reaching state-of-the-art accuracy in multilayer perceptrons (MLPs) and convolutional neural networks (CNNs). In this paper, we analyze the stability and similarity of predictions and neurons in MLPs and propose a new variation of one of the algorithms.
\end{abstract}

\section{Introduction}
When training neural networks there are many decisions that are made either manually or automatically through any of brute force, trial and error, or theoretical assumptions. These decisions include the overall architecture of the network, the learning rate, the initialization, and the learning algorithm among others. Each of these decisions results in a different model. Occasionally, as can be the case with different weight initialization techniques, the trained network converges to the same local extrema. 

Backpropagation is the most widely used algorithm to train a neural network[1]. The development of differential, non-linear, activation made it possible to use the chain rule to train deep neural networks. 

However, it is now believed that due to the weight transpose in the backward pass, this is not a biologically plausible algorithm [2]. This led to a significant increase in algorithms such as feedback alignment (FA) and direct feedback alignment (DFA) [3]. There is much research that many of these algorithms perform nearly as well in the MNIST and CIFAR10 tests when trained on multilayer perceptrons (MLPs) [2,3,4,6]. In this paper, we will examine not only the accuracy and stability of these algorithms on these tasks, but also the similarity of the predictions and neurons of models trained using these algorithms.

\section{Algorithms}

We will be comparing 5 different training algorithms. The first, backpropagation, is the benchmark we will be comparing everything to. The other four are variations of feedback alignment.

\subsection{Backpropagation}

By far the mostly widely used algorithm used to train neural networks. By backpropagating the gradient through each layer we can efficiently and effectively train deeper networks. The backprop algorithm is as follows,

\begin{align*}
    \delta W_i & = -((W^T_{i+1}\delta z_{i+1}) \odot \phi^{'}(z_i))y^T_{i-1} \\
    \delta z_{i+1} &= \diffp{\mathcal{L}}{{z_{i+1}}} \\
    y_i &= \phi(z_i) \\
    z_i &= W_i y_{i-1} + b_i
\end{align*}

Backpropagation tend to set the highest accuracy in MLPs, RNNS, CNNs, and ViTs. Newer algorithms attempt to achieve this accuracy with caveats of holding less information, not having a weight transport, or converging faster.

\subsection{Feedback Alignment}

In feedback alignment algorithms, the weight matrix W information in the back pass is replaced with a random matrix B. This allows us to not need a weight transport in the back pass. Shown in figure 1 below is a comparison of training in backpropagation and feedback alignment methods.

\begin{figure}[h!]
  \includegraphics[width=\textwidth]{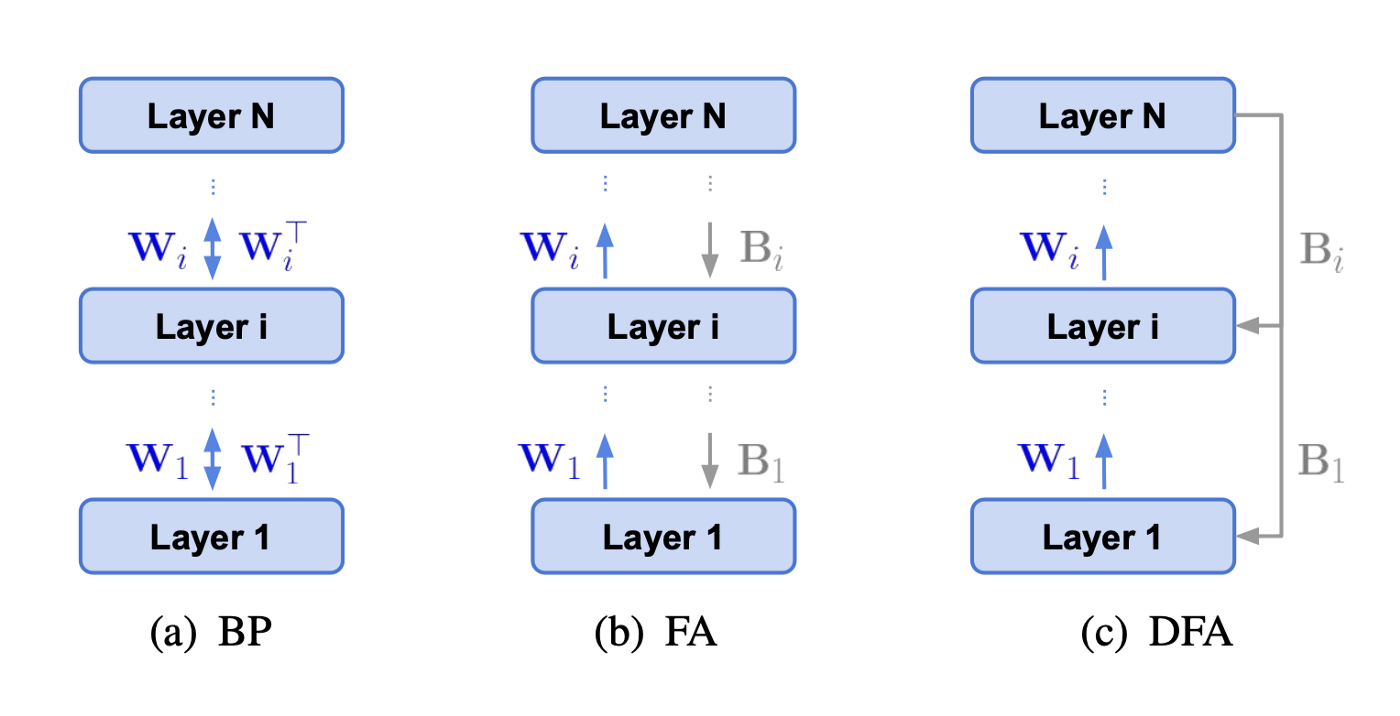}
    \caption{Comparison of training algorithms}
\end{figure}

Notice that in all feedback alignment algorithms a random matrix B is used in the backpass instead of the transport of the W matrix. Below are the formal definitions of the feedback alignment algorithms we will test,

\subsubsection{Vanilla Feedback Alignment}

Feedback alignment is an algorithm designed to bypass the $W^T$ weight transpose. Instead, a random matrix B with the same dimensions as $W^T$ for each layer in the network is created. Although this algorithm tends to learn a little slower in the early epochs, information is able to be learned through this random transformation. The algorithm is as follows,

\begin{align*}
    \delta W_i & = -((B_{i+1}\delta z_{i+1}) \odot \phi^{'}(z_i))y^T_{i-1} \\
    \delta z_{i+1} &= \diffp{\mathcal{L}}{{z_{i+1}}} \\
    y_i &= \phi(z_i) \\
    z_i &= W_i y_{i-1} + b_i
\end{align*}

\subsubsection{Uniform Signed Feedback Alignment}

Uniform signed feedback alignment is very similar to vanilla FA. However, unlike FA it updates the B matrix as follows, using some information from $W^T$. Specifically, the algorithm is,

\begin{align*}
    \delta W_i & = -((B_{i+1}\delta z_{i+1}) \odot \phi^{'}(z_i))y^T_{i-1} \\
    \delta z_{i+1} &= \diffp{\mathcal{L}}{{z_{i+1}}} \\
    y_i &= \phi(z_i) \\ 
    z_i &= W_i y_{i-1} + b_i
\end{align*}

And after every iteration we update the B matrix as follows,

\begin{align*}
    B_i & = \text{sign}(W_i^T) \\
\end{align*}

Therefore, this method is somewhat of a hybrid between the between backpropagation and vanilla FA. Unlike in vanilla FA, the B matrix is now a function of $W^T$. However, unlike backpropagation it is not a perfect representation on the weight transport as it only contains partial information on whether each element is non negative.

\subsubsection{Direct Feedback Alignment}

For direct feedback alignment (DFA) we directly propagate the error to each layer. Instead of $B$ having the same dimensions as $W^T$ , $B_i$ has the same input dimension as $W$ but the output dimension equal to the output layer

\begin{align*}
    \delta W_i & = -((B_{i}\delta z_{n}) \odot \phi^{'}(z_i))y^T_{i-1} \\
    \delta z_{n} &= \diffp{\mathcal{L}}{{z_{n}}} \\
    y_i &= \phi(z_i) \\
    z_i &= W_i y_{i-1} + b_i
\end{align*}

DFA was created as a means to effectively use feedback alignment in larger and deeper networks [5]. Whereas FA gradients may get lost during backwards pass through many randomized B layers DFA attempts to resolve that noise.

\subsubsection{Weighted Direct Feedback Alignment}

We propose a variation of DFA that slows down the learning rate of in the earlier layers by the
59 following logic,

\begin{align*}
    lr_j = lr \frac{n\sqrt{j}}{\sum_{i=1}^n \sqrt{i}}
\end{align*}

Notice that the average learning rate over all layers i to n is the starting learning rate $lr$. Therefore, since DFA is projecting the error propagation directly across all layers, we can slow down this change and avoid rapid changes to earlier layers using this method. DFA is expected to rival BP in deeper networks [6]. We propose that this change could increase stability in the training of these deeper networks.

\section{Procedure}

In order to compare these algorithms we build models with architecture as follows:

\subsection{MNIST Architecture}

We flattened the 28*28 images to have the following model architecture,

\begin{align*}
    \text{Layer}_{\text{input}}& : \text{Size} : (786, 768) . . . \text{Activation: tanh} \\
    \text{Layer}_{\text{hidden 1}}& : \text{Size} : (768, 256) . . . \text{Activation: tanh}\\
    \text{Layer}_{\text{hidden 2}}& : \text{Size} : (256, 128) . . . \text{Activation: tanh} \\
    \text{Layer}_{\text{output}}& : \text{Size} : (128, 10) . . . \text{Activation: sigmoid} \\
\end{align*}

\subsection{CIFAR10 Architecture}

We flattened the 3*32*32 images to have the following model architecture,

\begin{align*}
    \text{Layer}_{\text{input}}& : \text{Size} : (3072, 768) . . . \text{Activation: tanh} \\
    \text{Layer}_{\text{hidden 1}}& : \text{Size} : (768, 256) . . . \text{Activation: tanh}\\
    \text{Layer}_{\text{hidden 2}}& : \text{Size} : (256, 128) . . . \text{Activation: tanh} \\
    \text{Layer}_{\text{output}}& : \text{Size} : (128, 10) . . . \text{Activation: sigmoid} \\
\end{align*}

For each dataset we created 10 models of each type. For the $i$th model $i \in [1, 10]$, of each type we set the random seed equal to $i$ before initializing the weights of the model. That way, for each type of model, the $i$th model started as the exact same position. Therefore, any differences in the model are solely due to the differences in the training algorithm. All models were trained for 60 epochs with a learning rate of 1e-4. We used cross-entropy loss for all our models.

\section{Experimental Results}

\subsection{Accuracy Comparison}
79 For a similar comparison of the training process of these algorithms, all models were trained with a simple minor weight decay but without an optimizer. For accuracy purposes and for bench marking we included BP with an Adam optimizer.

\begin{table}[h]
  \caption{Accuracy Comparison}
  \label{sample-table}
  \centering
  \begin{tabular}{lll}
    \toprule
    \multicolumn{2}{c}{}                   \\
    \cmidrule(r){2-3}
    Algorithm     & MNIST     & CIFAR 10 \\
    \midrule
    BP & 0.973  & 0.491     \\
    DFA & 0.961  & 0.480    \\
    DFA Weighted & 0.960  & 0.478    \\
    FA     & 0.959 & 0.454     \\
    FA Uni Sign     & 0.955       & 0.471  \\
    \bottomrule
  \end{tabular}
\end{table}

Notice that while FA and Uniform Sign FA preform on par with BP, and DFA in MNIST, it suffers once more parameters and complexity is added. On the other hand, DFA tends to perform very well and stays in line, albeit a little below BP, for both tasks.

\subsection{Prediction Similarity}

\subsubsection{Stability of the algorithms}

First we will examine the similarity of the output of each model. We will look at the cosine similarity between the top predicted class from each of the 10 iterations for each model.

\begin{table}[h]
  \caption{Cosine Similarity of Predictions for each Algorithm}
  \label{sample-table}
  \centering
  \begin{tabular}{lll}
    \toprule
    \multicolumn{2}{c}{}                   \\
    \cmidrule(r){2-3}
    Algorithm     & MNIST     & CIFAR 10 \\
    \midrule
    BP & 0.98  & 0.851     \\
    DFA & 0.99  & 0.873    \\
    DFA Weighted & 0.99  & 0.877    \\
    FA     & 0.99 & 0.794     \\
    FA Uni Sign     & 0.99       & 0.825  \\
    \bottomrule
  \end{tabular}
\end{table}

From Table 2, we notice that from the MNIST results that when the model has few parameters the algorithms all tend to converge to similar results. This may be largely due to the fact that the MNIST accuracy is so high that the the few pictures the models are getting wrong are ambiguous and all the models fail to accurately classify them.

Comparing the CIFAR10 results, BP, DFA, and Weighted DFA tend to produce relatively stable models. Especially weighted DFA when the similarity scores among the predicted output are highest. This would suggest that Weighted DFA tends to be less dependent on the randomization of the B matrix then the other algorithms.

Lastly, we note that using the Uniform Sign FA, we tend to get back some of the stability lost using vanilla FA. Vanilla FA has the most randomized parameters and subsequently takes the lost to train and has the least stability.

\subsubsection{Comparison of Predictions Across Algorithms}

\begin{table}[h]
  \caption{Cosine Similarity Across Algorithm Outputs for CIFAR10}
  \label{sample-table}
  \centering
  \begin{tabular}{llllll}
    \toprule
    \multicolumn{2}{c}{}                   \\
         & BP     & DFA & DFA Weighted & FA & FA Uni Sign \\
    \midrule
    BP & -  & 0.844 & 0.853 & 0.806 & 0.831     \\
    DFA & 0.844  & - & 0.891 & 0.791 & 0.826     \\
    DFA Weighted & 0.853  & 0.891 & - & 0.791 & 0.828     \\
    FA     & 0.806  & 0.791 & 0.791 & - & 0.806     \\
    FA Uni Sign   & 0.831  & 0.826 & 0.828 & 0.806 & -     \\
    \bottomrule
  \end{tabular}
\end{table}

In Table 3, we record the average similarity scores of each instance of each model to the other instances. We see that our weighted DFA method most resembles the output of backpropagation followed by regular DFA. Much like, Table 2 we see that the more stable algorithms tend to have similar outputs compared to each other. BP is much closer to DFA methods than FA methods. 

As expected methods in the same type tend to be more similar to eachother. Noticeably, there is much less difference in Weighted DFA and DFA than there is between FA and Uniform Sign FA.

\subsection{Layer Similarity}

Lastly, we will measure the similarity among the 4 layers of the trained models. To do so we will once again measure the cosine similarity of the each of the layers both within the same algorithm and across algorithms.

Given the initial layer weights produced given the random seed, we then train a model starting with these weights using each of the five training algorithms. Then, for each seed and for each layer we check the cosine similarity between than layer for each of the 5 training algorithms. The results are as follows,

\begin{table}[h]
  \caption{Cosine Similarity of First Layer Weights Across Algorithms}
  \label{sample-table}
  \centering
  \begin{tabular}{llllll}
    \toprule
    \multicolumn{2}{c}{}                   \\
         & BP     & DFA & DFA Weighted & FA & FA Uni Sign \\
    \midrule
    BP & -  & 0.993 & 0.997 & 0.989 & 0.982     \\
    DFA & 0.993  & - & 0.997 & 0.982 & 0.976     \\
    DFA Weighted & 0.997  & 0.997 & - & 0.986 & 0.980     \\
    FA     & 0.989  & 0.982 & 0.986 & - & 0.973     \\
    FA Uni Sign   & 0.982  & 0.976 & 0.980 & 0.973 & -     \\
    \bottomrule
  \end{tabular}
\end{table}

In the first layer we can see that the most algorithms, are highly correlated. Weighted DFA tends to be the most similar to BP.

\begin{table}[h]
  \caption{Cosine Similarity of Third Layer Weights Across Algorithms}
  \label{sample-table}
  \centering
  \begin{tabular}{llllll}
    \toprule
    \multicolumn{2}{c}{}                   \\
         & BP     & DFA & DFA Weighted & FA & FA Uni Sign \\
    \midrule
    BP & -  & 0.855 & 0.863 & 0.926 & 0.971     \\
    DFA & 0.855  & - & 0.904 & 0.799 & 0.835     \\
    DFA Weighted & 0.863  & 0.904 & - & 0.807 & 0.843     \\
    FA     & 0.926  & 0.799 & 0.807 & - & 0.905     \\
    FA Uni Sign   & 0.971  & 0.835 & 0.843 & 0.905 & -     \\
    \bottomrule
  \end{tabular}
\end{table}

In the middle layers we can see that DFA starts to have some significant differences from BP and FA. Due to the direct propagation of the error term these methods are updated in a more distinct method than FA and BP.

\begin{table}[h]
  \caption{Cosine Similarity of Output Layer Weights Across Algorithms}
  \label{sample-table}
  \centering
  \begin{tabular}{llllll}
    \toprule
    \multicolumn{2}{c}{}                   \\
         & BP     & DFA & DFA Weighted & FA & FA Uni Sign \\
    \midrule
    BP & -  & 0.104 & 0.098 & 0.225 & 0.782     \\
    DFA & 0.104  & - & 0.857 & 0.062 & 0.107     \\
    DFA Weighted & 0.098  & 0.857 & - & 0.057 & 0.112     \\
    FA     & 0.225  & 0.062 & 0.057 & - & 0.257     \\
    FA Uni Sign   & 0.782  & 0.107 & 0.112 & 0.257 & -     \\
    \bottomrule
  \end{tabular}
\end{table}

But in the output layer shown in Table 6, the algorithms really start to have high cosine distances to each other. For most algorithms, this layer is pretty distinct. This layer is the only layer we see significant gaps between the DFA and Weighted DFA. Note that for this layer, the FA methods tends to be far more similar to BP than the DFA methods. This is especially true for Uni Sign FA.

\section{Conclusion}

After studying various training algorithms for MLPs, we see that in terms of similarity of weights across layers particularly in the earlier hidden layers, stability of algorithm, and similarity of final predictions, the Weighted DFA tends to best mimic the backpropagation algorithm. However, in terms of accuracy it tends to slightly underperform DFA on average.

In contrast, FA algorithms tend to update the first layers faster than DFA and BP and tend to be less stable. We also saw that FA fell off faster when we added more complexity to the problem when we shifted from MNIST to CIFAR10.

It remains to be seen if these similarities carry over to training of CNNs.Recent studies show that DFA performs well on CNNs in terms of accuracy [7].

It is clear that FA algorithms cannot handle increased complexity or deeper networks as efficiently as BP. However, both the Weighted DFA and regular DFA algorithms seem to be plausible candidates for moderate sized MLPs.

\section*{References}

{
\small

[1] Rumelhart, David E., Geoffrey E. Hinton, and Ronald J. Williams. Learning internal representations by error propagation. California Univ San Diego La Jolla Inst for Cognitive Science, 1985.

[2] Song, Yuhang, et al. "Can the Brain Do Backpropagation?—Exact Implementation of Backpropagation in Predictive Coding Networks." Advances in neural information processing systems 33 (2020): 22566-22579.

[3] Nøkland, Arild. "Direct feedback alignment provides learning in deep neural networks." Advances in neural information processing systems 29 (2016).

[4] Akrout, Mohamed, et al. "Deep learning without weight transport." Advances in neural information processing systems 32 (2019).

[5] Moskovitz, Theodore H., Ashok Litwin-Kumar, and L. F. Abbott. "Feedback alignment in deep convolutional networks." arXiv preprint arXiv:1812.06488 (2018).

[6] Launay, Julien, et al. "Direct feedback alignment scales to modern deep learning tasks and architectures." Advances in neural information processing systems 33 (2020): 9346-9360.

[7] Han, Donghyeon, and Hoi-jun Yoo. "Efficient convolutional neural network training with direct feedback alignment." arXiv preprint arXiv:1901.01986 (2019).

}

\end{document}